\documentclass[letterpaper, 10 pt, conference]{ieeeconf}  

\IEEEoverridecommandlockouts      

\usepackage{amsmath} 
\usepackage{graphicx}
\usepackage{booktabs}
\usepackage{amssymb}
\usepackage{siunitx}
\usepackage[english]{babel}
\usepackage[noadjust]{cite}
\usepackage{units}
\usepackage{epstopdf}
\usepackage[font=small]{caption}
\usepackage{subcaption}
\usepackage{comment}
\usepackage{amsfonts}
\usepackage{bm}
\usepackage{algorithm}

\usepackage{amsthm}
\usepackage{algorithmic}

\usepackage{color}

\usepackage{xcolor}
\newcommand{\remove}[1]{}

\DeclareMathOperator*{\argmin}{arg\,min}

\theoremstyle{plain}

\newtheorem{theorem}{Theorem}

\newtheorem{definition}[theorem]{Definition}

\title{\LARGE \bf
Adaptive Safe Merging Control for Heterogeneous Autonomous Vehicles using Parametric Control Barrier Functions 
}

\author{Yiwei Lyu$^1$, Wenhao Luo$^2$ and John M. Dolan$^2$
\thanks{$^*$This work was supported by the CMU Argo AI Center for Autonomous
Vehicle Research.}
\thanks{$^1$The author is with the Department of Electrical and Computer Engineering, Carnegie Mellon University, Pittsburgh, PA, 15213 USA. Email: {\tt \small yiweilyu@andrew.cmu.edu}}%
\thanks{$^{2}$The authors are with the Robotics Institute, Carnegie Mellon University, Pittsburgh, PA 15213 USA. Email: {\tt \small \{wenhaol, jmd\}@cs.cmu.edu}}%
}

\begin{document}

\maketitle
\thispagestyle{empty}
\pagestyle{empty}

\begin{abstract}
 
With the increasing emphasis on the safe autonomy for robots, model-based safe control approaches such as Control Barrier Functions have been extensively studied to ensure guaranteed safety during inter-robot interactions. In this paper, we introduce the Parametric Control Barrier Function (Parametric-CBF), a novel variant of the traditional Control Barrier Function to extend its expressivity in describing different safe behaviors among heterogeneous robots. 
Instead of assuming cooperative and homogeneous robots using the same safe controllers, the ego robot is able to model the neighboring robots' underlying safe controllers through different Parametric-CBFs with observed data. Given learned parametric-CBF and proved forward invariance, it provides greater flexibility for the ego robot to better coordinate with other heterogeneous robots with improved efficiency while enjoying formally provable safety guarantees. We demonstrate the usage of Parametric-CBF in behavior prediction and adaptive safe control in the ramp merging scenario from the applications of autonomous driving. 
Compared to traditional CBF, Parametric-CBF has the advantage of capturing varying drivers' characteristics given richer description of robot behavior in the context of safe control. Numerical simulations are given to validate the effectiveness of the proposed method.

\end{abstract}

\section{Introduction}

Although there have been strong strides in the area of heterogeneous robot control, how to obtain theoretically guaranteed safety in heterogeneous robots interaction is still an open challenge. Among the model-based safe control approaches, Control Barrier Function (CBF) is a popular choice due to its forward invariant property in enforcing persistent safety. In recent progress on safe design for autonomous robots using CBF \cite{Ames2014,ames2019control,wang2017tor,zeng2020safetycritical,lyu2021probabilistic}, surrounding robots are often assumed to be fully cooperative or passively moving with constant velocity in the collision avoidance scenario. However, in a more realistic setting when robots operate in an unknown environment and interact with other autonomous systems, e.g. other unknown autonomous driving cars, it is desired for the robots to take observations and infer the underlying safe behavior strategy for the other robots, so that the ego robot could behave more safely and efficiently during such interactions.

When adopting safety-critical control into the domain of autonomous driving, new challenges arise. Considering the fact that autonomous vehicles will have to share roads with human drivers for a very long time, how to improve the expressivity of autonomous vehicles' safe-critical controllers, so that the generated behavior can be better understood by human drivers, remains an open problem. As a milestone towards the final goal of mixed human-robot autonomy, it is desired to explore methods to help heterogeneous autonomous vehicles to achieve richer behavior description and characterization while safely interacting with each other.

\begin{figure}
    \centering
    \includegraphics[width = 0.7\linewidth]{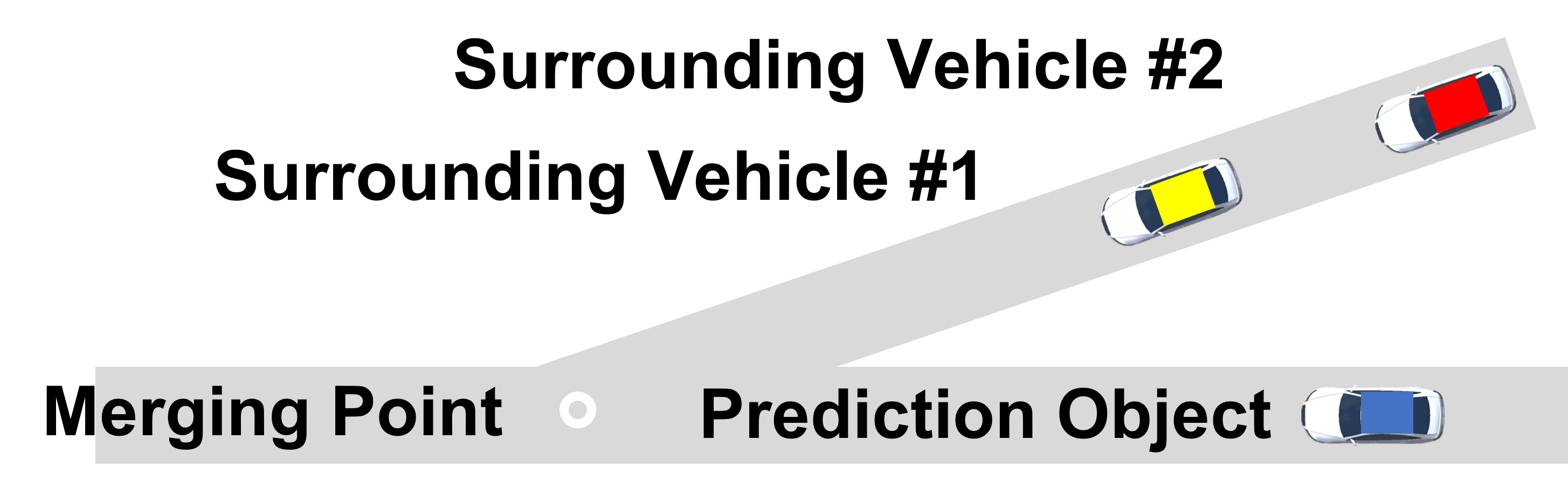}
    \caption{\footnotesize
    \label{pred_scenario}
   Prediction Scenario. The blue car on the main road is the prediction object whose driving style needs to be estimated. The yellow car and the red car are the surrounding vehicles on the ramp. All three vehicles are using Parametric-CBF-based controllers and are interacting with each other.}
\end{figure}

Motivated by these considerations, we focus on the learning and safe design for interaction of heterogeneous autonomous vehicles in ramp merging scenario. This paper extends our previous work on CBF-based safety-assured merging control in \cite{lyu2021probabilistic} 
and presents the following \textbf{contributions}: 1) We propose the novel idea of Parametric-CBF, a variant of traditional CBF that gives a richer behavior description and preserves the property of rendering a forward invariant safe set for the robots;
2) We present a novel safe adaptive merging algorithm that integrates the safe behavior prediction of heterogeneous robots and safe control for the ego robot using the learned parameters of the Parametric CBF, yielding improved task efficiency for the ego robot with safety guarantee.
3) We demonstrate the effectiveness of the Parametric-CBF based behavior predication and safe control through experimental results in the task of ramp merging in the autonomous driving domain; 
Our mechanism enables the robot to model the behavior of other entities first and take appropriate actions accordingly, which makes it generally applicable to other robotics applications.

\section{Related Work}

Safe control in terms of collision avoidance is critical for designing interactive robot behaviors. To minimize the deviation from robots' primary task execution due to safety consideration, reactive collision avoidance methods such as reciprocal velocity obstacles (RVO) \cite{van2008reciprocal, alonso2012reciprocal}, 
safety barrier certificates \cite{wang2017tor, wang2017safe, luo2020multi} and buffered Voronoi cells \cite{wang2019game, angeris2019fast} have been presented to minimally revise the robot's task-related controller subject to collision avoidance constraints. The collision-free motion often relies on the key assumption of reciprocal or passive behaviors, where each robot is assumed to either employ the homogeneous safe controller (fully cooperative) or move with constant velocity (non-cooperative). When a robot is interacting with unknown agents that employ different behavior design, it is necessary to leverage the observations on the other agents to build effective models for their behaviors and achieve reliable safe interaction between the robots.

On the other hand, predicting behavior patterns of interactive heterogeneous multi-robot systems is often modeled as parameter identification problem. \cite{grover2020parameter} presents a method to estimate the system dynamics parameter of a robot using an optimization-based controller, assuming that the CBF-based safety constraint of the optimization-based controller is known. This would require foreknowledge of what kind of safe controller the robot is using, and therefore, the same identification framework cannot be applied directly to other robots with different safe controller parameters. \cite{wang2016acc} presents a decentralized control framework for heterogeneous multi-robot interaction by assuming the same reciprocal behaviors but with different mobility capability to identify, and the causality of the difference in generated safe behavior is simplified as the difference in robots' acceleration limits.
To capture different safe behaviors produced by heterogeneous safe constraints, it is desired to directly characterize and infer the behavior pattern modelled by learned safe constraints, e.g. CBF-based control constraints, so that the safe behavior of unknown agents could be better predicted.

There have been some recent efforts to construct variants of CBFs with parameter identification to describe various controllers. \cite{jagtap2020compositional,jagtap2020formal} introduced techniques to search for parametric CBFs to synthesize controllers for optimal control, but the focus was on the parameter computation of an assumed particular form of CBF, rather than providing a formal definition of parametric CBF that can be applied generally. \cite{djaballah2017construction} introduced a systematic way to construct parametric barrier functions using interval analysis. However, the heavy computation associated with solving the complicated formulation of the parameterized CBFs, which varies case-by-case, prohibits the applicability to large-scale multi-robot systems in real time.
In this paper, we propose a general formulation of the Parametric-CBF 
to leverage a linear combination of candidate CBFs characterizing various risk tolerance levels when the autonomous system is approaching the safety set boundary, therefore enabling distinct and richer behavior expressions for efficient learning and safe control.

\section{Background on Control Barrier Functions}

Control Barrier Functions (CBF) \cite{ames2019control} are used to define an admissible control space for safety assurance of dynamical systems.
One of its important properties is its forward-invariance guarantee of a desired safety set. Consider the following nonlinear system in control affine form:
\begin{equation}\label{eq:nonlinear}
    \dot x = f(x)+g(x)u
\end{equation}
where $x\in \mathcal{X}\subset \mathbb{R}^n$ and $u\in\mathcal{U}\subset \mathbb{R}^m$ are the system state and control input with $f$ and $g$ assumed to be locally Lipschitz continuous.
A desired safety set $x\in\mathcal{H}$ can be denoted by a safety function $h(x)$: 
\begin{equation}\label{eq:safeset_general}
\mathcal{H} =\{x \in \mathbb{R}^n : h(x)\geq 0\}
\end{equation}
Thus the control barrier function for the system to remain in the safety set can be defined as follows \cite{ames2019control}:
\begin{definition}
(Control Barrier Function) Given a dynamical system (\ref{eq:nonlinear}) and the set $\mathcal{H}$ defined in (\ref{eq:safeset_general}) with a continuously differentiable function $h:\mathbb{R}^n\rightarrow \mathbb{R}$, then $h$ is a control barrier function (CBF) if there exists a class $\mathcal{K}$ function for all $x\in \mathcal{X}$ such that 
\begin{equation}\label{eq:cbf_def}
    \sup_{u\in\mathcal{U}} \ \{L_f h(x)+L_g h(x) u\}\geq -\kappa \big(h(x)\big)
\end{equation}
\end{definition}
\noindent
where $\dot{h}(x,u)=L_f h(x)+L_g h(x) u$ with $L_f h, L_g h$ as the Lie derivatives of $h$ along the vector fields $f$ and $g$.

A commonly selected class $\mathcal{K}$ function is $\kappa (h(x))=\gamma h(x)$~\cite{ames2019control,zeng2020safetycritical,he2021rulebased}, where $\gamma\in\mathbb{R}^{\geq 0}$ is a CBF design parameter controlling system behaviors near the boundary of $h(x)=0$. Hence, the admissible control space in (\ref{eq:cbf_def}) can be redefined as 
\begin{equation}\label{eq:cbf}
    \mathcal{B}(x)=\{u\in\mathcal{U}:\dot{h}(x,u) + \gamma h(x)\geq 0\; \}
\end{equation}
It is proved in \cite{ames2019control} that any controller $u\in\mathcal{B}(x)$ will render the safe state set $\mathcal{H}$ forward-invariant, i.e., if the system (\ref{eq:nonlinear}) starts inside the set $\mathcal{H}$ with $x(t=0)\in \mathcal{H}$, then it implies $x(t)\in\mathcal{H}$ for all $t>0$ under controller $u\in\mathcal{B}(x)$.
However, 
this particular form as $\kappa (h(x))=\gamma h(x)$ is limited in describing complicated system behaviors when approaching to the boundary of $h(x)=0$, so does the other particular form as $\kappa (h(x))=\gamma h^3(x)$ used in \cite{wang2016acc}. Thus a more general form capturing a richer nonlinear behavior descriptions is needed.

\section{Method}
\label{method}

In this section, we start by introducing the formal definition of Parametric-CBF followed with the properties proof. Then Parametric-CBF based safe controller design and Parametric-CBF based behavior style prediction are introduced. Integrating the two tasks together, the safe adaptive merging control algorithm is presented. Finally, evaluation on advantages of Parametric-CBF is performed.

\subsection{Formal Definition of Parametric-CBF}

The main difference between the proposed Parametric-CBF and traditional CBF is the different choice of the mapping function $\kappa (h(x))$. 
\begin{definition}
(Parametric-Control Barrier Function) Given a dynamical system (\ref{eq:nonlinear}) and the set $\mathcal{H}$ defined in (\ref{eq:safeset_general}) with a continuously differentiable function $h:\mathbb{R}^n\rightarrow \mathbb{R}$, then $h$ is a Parametric-Control Barrier Function (Parametric-CBF) for all $x\in \mathcal{X}$ such that 
\begin{equation}\label{eq:parametric_cbf_def}
    \sup_{u\in\mathcal{U}} \ \{\dot{h}(x,u)\}\geq -\alpha H(x)
\end{equation}
where parameter vector $\alpha = \begin{bmatrix}
\alpha_1 & \alpha_2 & \alpha_3 & \dots &\alpha_q \end{bmatrix}\in\mathbb{R}^q$ with $\forall\alpha_p\in\mathbb{R}^{\geq 0}$ for $p\in[q]$. $H(x) = \begin{bmatrix}
h(x) & h^3(x) & h^5(x) & \dots & h^{2q-1}(x) \end{bmatrix}^T$, $q\in \mathcal{N}$.
\end{definition}
In Parametric-CBF, $\kappa (h(x))$ is constructed as a polynomial function $\kappa (h(x)):=\alpha_1 h(x)+\alpha_2 h^3(x) + \alpha_3 h^5(x)+\dots +\alpha_q h^{2q-1}(x)$ with $H(x)$ as a set of basis functions containing independent odd-powered power functions $h^{2q-1}(x)$. These odd-powered power functions themselves are candidate class $\mathcal{K}$ functions \cite{wang2017tor}. The
intuition behind the polynomial design is that the relative weighting of
its basis, composed of different-order component functions, regulates how fast the states of the system can
approach the boundary of the safe set, and therefore how well the
polynomial function can capture the system’s behavior near the
boundary. The number of basis functions $q$ could be empirically pre-determined by user depending on the degrees of flexibility as needed.
Note that the proposed Parametric-CBF is a more general formulation that includes all the $\kappa (h(x))$ choices used in \cite{ames2019control,wang2016acc,wang2017tor,zeng2020safetycritical,he2021rulebased}.

\subsection{Properties Proof of Parametric-CBF}

In this section we prove that 
function $\kappa(h(x))= \alpha H(x)$ is a class $\mathcal{K}$ function whose definition is as follows \cite{wang2016acc}:
\begin{definition}
A continuous function $\beta: [0,a)\rightarrow [0,\infty)$ for some $a$ is called a class $\mathcal{K}$ function if 1) it is strictly increasing and 2) $\kappa(0)=0$.
\end{definition}
To verify the validity of the proposed Parametric-CBF, we need to prove that the function $\kappa(h(x))= \alpha H(x)$ has the properties of strictly increasing and passing the origin.
\begin{proof}

\textbf{Strictly increasing}: The easiest way to prove this statement is to calculate the first-order derivative of $\kappa(\cdot)$ w.r.t $h$:
\begin{equation}
    \begin{split}
        {\frac{\partial\kappa}{\partial h}} = \frac{\partial}{\partial h}(\alpha_1 h(x)+\alpha_2 h^3(x) +\dots +\alpha_q h^{2q-1}(x)) \\
        = \alpha_1 + 3\alpha_2 h^2(x) + \dots + (2q-1)\alpha_q h^{2q-2}(x) 
    \end{split}
\end{equation}
With non-negative parameters $\alpha_1,\ldots,\alpha_q$ and the even-powered power functions $h^2,\ldots,h^{2q-2}$, it is straightforward that ${\frac{\partial\kappa}{\partial h}}\geq 0$ for $\forall h\geq 0$ and ${\frac{\partial\kappa}{\partial h}}> 0$ for $\forall h> 0$, indicating that $\kappa(\cdot)$ is a strictly increasing function.

\textbf{Passes the origin}: Substituting $h(x) = 0$ into the polynomial function $\alpha H(x)$, we have 
\begin{equation}
    \begin{split}
         H(x)=\begin{bmatrix}
h(x) \\ h^3(x) \\ h^5(x) \\ \vdots \\ h^{2q-1}(x) \end{bmatrix} 
= \begin{bmatrix}
0 \\ 0 \\ 0 \\ \vdots \\ 0 \end{bmatrix} \implies \kappa(0)=0\cdot\alpha=0
    \end{split}
\end{equation}
Therefore, the constructed polynomial function $\kappa(h(x))=\alpha H(x)$ is proved to be a class $\mathcal{K}$ function. 
\end{proof}

Hence, the admissible control space (\ref{eq:parametric_cbf_def}) can be redefined as 
\begin{equation}\label{eq:cbf}
    \mathcal{C}(x)=\{u\in\mathcal{U}:\dot{h}(x,u) + \alpha H(x)\geq 0\; \}
\end{equation}
and with our $\kappa(\cdot)$ being a class $\mathcal{K}$ function, it is proved in \cite{ames2019control} that any controller $u\in\mathcal{C}(x)$ will render the safe state set $\mathcal{H}$ forward-invariant using the comparison lemma, i.e., if the system (\ref{eq:nonlinear}) starts inside the set $\mathcal{H}$ with $x(t=0)\in \mathcal{H}$, then it implies $x(t)\in\mathcal{H}$ for all $t>0$ under controller $u\in\mathcal{C}(x)$.
Now we conclude the properties proof of Parametric-CBF.

\subsection{Parametric-CBF based Safe Controller}

In this work, the system dynamics of a vehicle can be described by the same double integrators as in \cite{lyu2021probabilistic}, since acceleration plays a key role in the safety considerations: 
\begin{equation} 
\begin{split}
    \dot{X} &=\begin{bmatrix}
    \dot{x}\\
    \dot{v}
    \end{bmatrix}
    =\begin{bmatrix}
    0_{2\times2}\; I_{2\times2}\\
    0_{2\times 2} \;0_{2\times 2}
    \end{bmatrix}
    \begin{bmatrix}
    x \\
    v
    \end{bmatrix}
    + \begin{bmatrix}
    0_{2\times2}\; 0_{2\times2}\\
    I_{2\times2}\; 0_{2\times2}
    \end{bmatrix}\begin{bmatrix}
    u\\ 0
    \end{bmatrix} \\
\end{split}
\label{dynamics}
\end{equation}
where $x\in\mathcal{X}\subset\mathbb{R}^2,v\in \mathbb{R}^2$ are the position and linear velocity of each car respectively and $u\in\mathbb{R}^2$ represents the acceleration control input. 
The safe controller is formulated as a quadratic program for heterogeneous multi-vehicles with the control input $u_i$.
\begin{equation}
\begin{split}
      &\min_{u_i\in\mathcal{U}_i} || u_i-\Bar{u}_i||^2 \\
    s.t \quad & {U}_i^{min} \leq  u_i \leq {U}_i^{max} \\
     &\dot{h}_{ij}(x, u) +\alpha_i H_{ij}(x)\geq 0, 
     \alpha_i = \begin{bmatrix}
\alpha_{i,1} & \ldots &\alpha_{i,q} \end{bmatrix}\\
H_{ij}(x) = &\begin{bmatrix}
h_{ij}(x) & h_{ij}^3(x) & h_{ij}^5(x) & \dots & h_{ij}^{2q-1}(x) \end{bmatrix}^T,  q\in \mathcal{N}
\end{split}
\label{qp}
\end{equation}
where $i,j$ are the indices of the pairwise vehicles. 
$\Bar{u}_i$ is the nominal expected acceleration for the ego vehicle $i$ to follow, and $U_i^{max}$ and $U_i^{min}$ are the ego vehicle's maximum and minimum allowed acceleration. We assume $\Bar{u}_i$ is computed by a higher-level task-related planner, for example, a behavior planner. We consider the particular choice of pairwise vehicle safety function $h_{ij}(x)$ and safety set $\mathcal{H}_i$ as follows.
\begin{equation}\label{eq:multi_safety}
\begin{split}
    \mathcal{H}(x)&=\{x\in\mathcal{X}: \; h_{ij}(x) = ||x_i-x_j||^2-R_{safe}^2\geq 0,\forall i\neq j\} 
\end{split}
\end{equation}
where $x_i,x_j$ are the positions of each pairwise set of vehicles and $R_{safe}\in\mathbb{R}^+$ is the minimum allowed safety distance.

\subsection{Heterogeneous Robots Behavior Prediction through Parameter Learning}
We assume each heterogeneous robot carries the safe controller (\ref{qp}) with different parameters $\alpha_i$ reflecting their various safe control behaviors, e.g. how aggressive they are in engaging collision avoidance scenario. Here we consider the behavior prediction task for ego vehicle $i$ over prediction object $j$, who is interacting with its surround vehicle $k$. To that end, the ego vehicle is able to observe the behavior of the prediction object $j$ and obtain the interactive dataset $\mathbb{D}^t=$\{$\dot{h}_{jk}^t,h^t_{jk}\}^{m}_{t=1}$ calculated from the observations during $m$ time steps (the position and velocity of the prediction object and its surrounding vehicles). $\dot{h}_{jk}^t$ denotes the change rate of ${h}_{jk}^t$ 
while $j$ assumes that surrounding vehicles $k$ are moving with piece-wise constant velocity sensed in real time, which is commonly assumed in many existing works on multi-agent collision avoidance, e.g. \cite{luo2020multi, zhu2019chance}.

Given that, the ego robot $i$ could perform ridge linear regression to find estimated $\bar{\alpha}_j$ for the prediction object $j$ as follows.
\begin{align}\label{eq:linear_regression}
\bar{\alpha}_j = \argmin_{\alpha_j} \sum_{t=1}^m \left\| \dot{h}_{jk}^t - \alpha_j {H}_{jk}^t  \right\|_2^2  + r \|\alpha_j\|_F^2
\end{align} 
where $r$ is a regularizer parameter and $\|\alpha_j\|_F$ is the Frobenius norm of the estimate parameter $\alpha_j$.

\subsection{Safe Adaptive Merging control Algorithm using Parametric-CBF}

Assuming that the learnt parameter $\alpha$ is correct, in this section it is proved that interactions between heterogeneous robots employing Parametric-CBF based safe controllers are always safe. The intuition behind the robots interaction mechanism design is that, robots have assumptions about other robots' movement when planning their own control, and this assumption can be arbitrary acceleration profiling. For the sake of simplicity, the common assumption of piece-wise constant velocity, made by many existing works in collision avoidance \cite{luo2020multi, zhu2019chance}, is chosen.

\begin{theorem}
For two interacting heterogeneous robots $i$ and $j$ employing Parametric-CBF based safe controllers, assuming 1) $j$ assumes that $i$ moves in piece-wise constant velocity, and 2) $i$ is aware of $j$'s assumption, then safety between pairwise robots $i,j$ can be formally guaranteed, regardless of whether $i$ acts as $j$ assumed.

\end{theorem}

\begin{proof}

\textbf{For ego robot $i$}, the safety constraint at timestep $t$ is derived as:
\begin{equation}
    \begin{split}
        h_{ij} &= ||x_i^t-x^t_j||^2-R_{safe}^2\\
        \dot{h}_{ij} &=2(x^t_i-x^t_j)^T(\dot{x}^t_i-\dot{x}^t_j)= 2{\Delta x_{ij}^t}^T\Delta v_{ij}^t\\
        & + 2{\Delta x_{ij}^t}^T u^t_i\Delta t - 2{\Delta x_{ij}^t}^T u_j^t \Delta t \geq -\alpha_i^T H_{ij}(x)
    \end{split}
    \label{i-constraint}
\end{equation}

\textbf{For the other robot $j$}, assuming $i$ is moving in piece-wise constant velocity sensed in real time with $u_i=0$, the safety constraint is similarly derived as:
\begin{equation}
    \begin{split}
        \dot{h}_{ji} &=2(x^t_j-x^t_i)^T(\dot{x}^t_j-\dot{x}^t_i)= 2{\Delta x_{ji}^t}^T\Delta v_{ji}^t + 2{\Delta x_{ji}^t}^T u^t_j\Delta t \\
        & = 2{\Delta x_{ij}^t}^T\Delta v_{ij}^t
         - 2{\Delta x_{ij}^t}^T u^t_j\Delta t \geq -\alpha_j^T H_{ji}(x)
    \end{split}
    \label{j-constraint}
\end{equation}

To make the two inequality constraints directly comparable, we add $\alpha_i^T H_{ij}(x)-\alpha_j^T H_{ij}(x)$ to both sides of Eq. \ref{i-constraint}, and we get: $2{\Delta x_{ij}^t}^T\Delta v_{ij}^t + 2{\Delta x_{ij}^t}^T u^t_i\Delta t - 2{\Delta x_{ij}^t}^T u_j^t \Delta t +\alpha_i^T H_{ij}(x) - \alpha_j^T H_{ij}(x) \geq -\alpha_j^T H_{ij}(x)$. Now compared to Eq. \ref{j-constraint}, we can further derive a new inequality constraint:
\begin{equation}
     2{\Delta x_{ij}^t}^T u_i^t \Delta t +(\alpha_i - \alpha_j)^T H_{ij}(x) \geq 0
     \label{equivalent-safety}
\end{equation}
This inequality constraint Eq. \ref{equivalent-safety} is therefore the sufficient condition to ensure that, as long as $u_i$, $\alpha_i$ and $\alpha_j$ satisfies this state-dependent inequality, the safety between pairwise interactive robots is formally guaranteed.
\end{proof}

\begin{algorithm}\footnotesize
\caption{Safe Adaptive Merging Algorithm using Parametric-CBF}
\begin{algorithmic}
\REQUIRE $\Delta x_{ij}, \Delta x_{jk}, \Delta v_{ij}, \Delta v_{jk}, \Delta t, R_{safe}$
\ENSURE $\bar{\alpha}_j,u_i$
\FOR{$t=1:m$} 
\IF{$\bar{\alpha}_j$ not converged} 
\STATE $h_{jk}^t(x) = ||x_j^t-x_k^t||^2-R_{safe}^2\geq 0$
\STATE calculation of $\dot{h}^t_{jk}$
\STATE add \{$\dot{h}_{jk}^t,h^t_{jk}$\} to dataset $\mathbb{D}^t$
\STATE learn Parametric-CBF parameter $\bar{\alpha}_j$ from
$\mathbb{D}^t$  (\ref{eq:linear_regression})
\ENDIF
\ENDFOR
\STATE choose the appropriate $\alpha_i$ based on $\bar{\alpha}_j$ 
\FOR{$t=m:N$} 
\STATE compute safety constraint parameter $A_{ij}^t,b_{ij}^t$
\STATE$u_i^t = \argmin_{u_i} ||u_i-\Bar{u}_i||^2$ with constraints in (\ref{qp})
\ENDFOR
\end{algorithmic}
\end{algorithm}

The framework of the proposed safe adaptive merging algorithm using Parametric-CBF is presented in Algorithm 1. $i,j,k$ are the indexes of the ego vehicle, the prediction object, and the interactive surrounding vehicle of the prediction object. $\Delta x_{ij}, \Delta x_{jk}$ are position differences, $\Delta v_{ij}, \Delta v_{jk}$ are velocity differences, and $\Delta t$ is the time interval. The two $for$ loops correspond to the time period of the prediction task and the control task respectively. It is assumed that the observation period is long enough to achieve a converged prediction result. Once the prediction of $\alpha_j$ is achieved, the appropriate $\alpha_i$ of the ego vehicle will be chosen accordingly subject to the constraint Eq. \ref{equivalent-safety}. The general rule is to resolve potential conflicts in both parities' driving strategies to avoid deadlock situation. If the prediction object $j$ is using a very aggressive driving style, the ego vehicle $i$ is set to be relatively conservative, and vice versa. In the second loop, the safety constraint parameter $A^t_{ij}$ and $b^t_{ij}$ are state-depedent variables, referring to the Parametric-CBF constraint in (\ref{qp}) which can be reorganized and written in the form of $A^t_{ij} u_i \leq b^t_{ij}$. By integrating Parametric-CBF based prediction and control together, we are able to achieve an adaptive merging control while accounting for formally provable safety.

\subsection{Evaluation of Parametric-CBF}

\subsubsection{\textbf{Richer Descriptive Information}}

The biggest advantage of Parametric-CBF is that, compared to traditional CBF, Parametric-CBF can describe more diverse autonomous system behavior patterns. With the polynomial function $\kappa(h(x))=\alpha H(x)$, Parametric-CBF can capture the system behavior characteristics to differing degrees. It especially benefits from the design of the linear combination of parameter vector $\alpha$ and high-order safety measurement vector $H(x)$, which re-formulate the nonlinear mapping relationship to be linear by changing the basis $h(x)$ in traditional CBF to $H(x)$.  

\subsubsection{\textbf{Broader Applications}} 

As introduced in the earlier section, Parametric-CBF can be simplified as $\dot{h}(x) = \alpha H(x)$, which makes it more generally applicable to various kinds of application scenarios, instead of limited to control tasks. One example is using machine learning approaches, e.g., ridge linear regression, to learn and predict the behavior of an autonomous system using a Parametric-CBF-based controller. More detailed examples will be provided in the following simulation section. Benefiting from this feature, the application of Parametric-CBF can be extended to different domains and has a broader impact.

\section{Applications}
\label{experiment}

\subsection{Prediction of Driving Style}

Different from the symmetric setting where all robots share the same CBF parameters as in \cite{borrmann2015control}, in the context of heterogeneous vehicles, vehicles do not have the same responsive behavior to potential collisions. Therefore, it is important to figure out how much responsibility the other vehicle would take for avoiding collision during interaction, which calls for the need of driving style prediction. The interactive driving scenario is shown in Fig. \ref{pred_scenario}. The task here is to predict the driving style, i.e., the Parametric-CBF parameter vector $\alpha$, of the prediction object, the blue car on the main road, while it's interacting with its surrounding vehicle. Heterogeneous robots' safety constraints are assumed to be active during the interactions. 

The prediction task is validated on 30 trials with randomly generated driving styles for the prediction object. Note, to demonstrate the formulation generality of proposed Parametric-CBF, the aforementioned particular choices of $\kappa(h(x))$ used in other papers \cite{ames2019control} ($\lambda h(x)$), \cite{wang2016acc} ($h^{3}(x)$), are also included in the tests. All the trials return converged prediction to ground truth value with an average root mean square error of $6.32\times10^{-6}$. It is observed that for the special cases of Parametric-CBF~\cite{ames2019control,wang2016acc}, the Parametric-CBF-based prediction framework does not lose its generality by achieving a consistent prediction performance. 

\begin{figure}
    \centering
    \includegraphics[width = 0.8\linewidth]{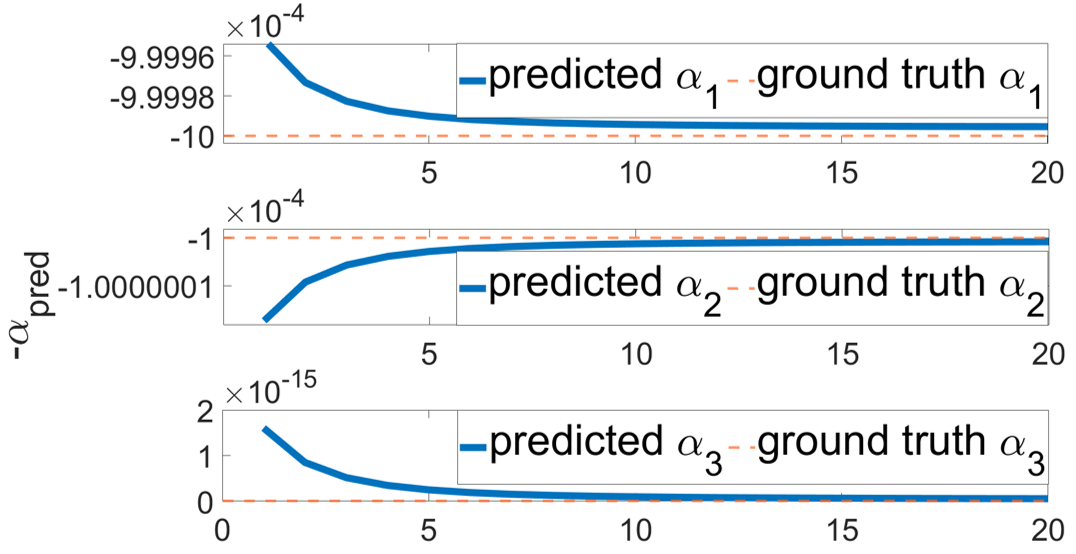}
    \caption{\footnotesize
    \label{prediction_demo}
   Learnt driving style $\alpha$ of the prediction object. The orange dashed lines are the ground truth of the parameter vector $\alpha$, and the blue lines are the learned parameter vector $\alpha$ over time.}
\end{figure}

Computational efficiency is also an advantage of Parametric-CBF. A prediction example is provided in Fig. \ref{prediction_demo}. The prediction results converges over time as more data points are collected and the learned parameter vector $\alpha$ reaches convergence in only 10 time steps, which is 0.1s, indicating it is computationally efficient enough to be applied in real-time applications.

\subsection{Adaptive Behavior in Control}
In this section, the purpose is to demonstrate the advantage of Parametric-CBF over traditional CBF in terms of behavior description richness in the safe control task. A two-vehicle interactive driving scenario is considered here, where both the ego vehicle on the main road and the other vehicle on the ramp use Parametric-CBF based safe controllers. The other vehicle's parameter $\alpha$ remains the same, meaning the driving style is kept the same in all the trials, and the construction of the ego vehicle's Parametric-CBF is varied.

\begin{figure}
    \centering
    \includegraphics[width = 0.7\linewidth]{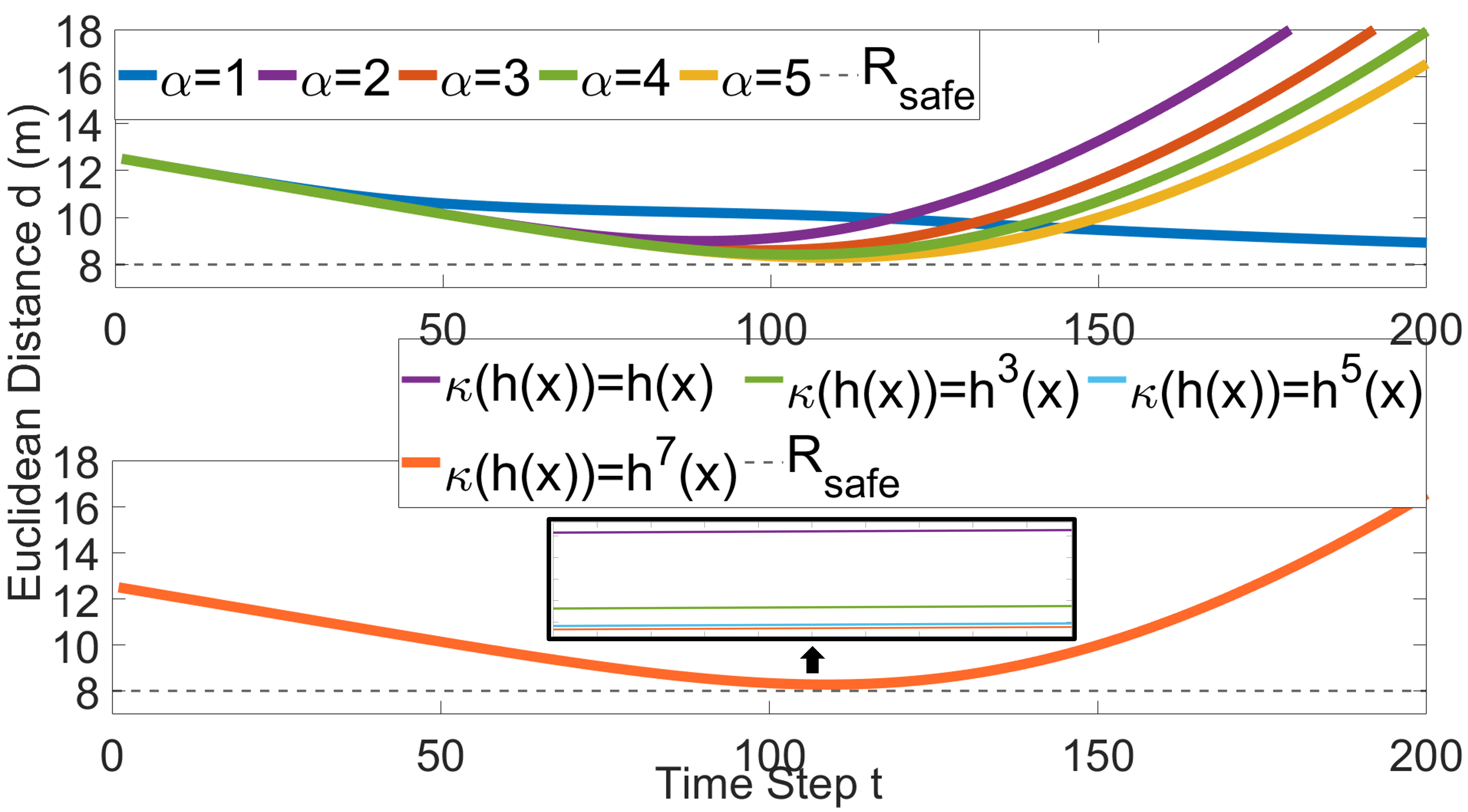}
    \caption{\footnotesize
    \label{effect_parameter_order}
   Performance of traditional CBF with particular forms of $\kappa(h(x))$~\cite{ames2019control,wang2016acc}: with different parameters (upper plot) and different orders(lower plot). The figures plot the inter-vehicle distance over time. The black dashed line indicates the predefined safety margin $R_{safe}$.}
\end{figure}

First, the performance of traditional CBF with various configurations of $\kappa(h(x))$ is shown in Fig. \ref{effect_parameter_order}. Note that the two different kinds of curvatures correspond to two merging results: in front of the other vehicle or in the back.  It is observed that, for traditional CBF, the larger Class $\mathcal{K}$ parameter is, the higher order the Class $\mathcal{K}$ function is, the more aggressive the driving behavior will be. Higher order class $\mathcal{K}$ function theoretically represents stronger tolerance of system movement towards the boundary of the safe set, however due to the simple composition of traditional CBF, the resulting behavior difference is too minor to tell in the lower plot.

\begin{figure}
    \centering
    \includegraphics[width = 0.8\linewidth]{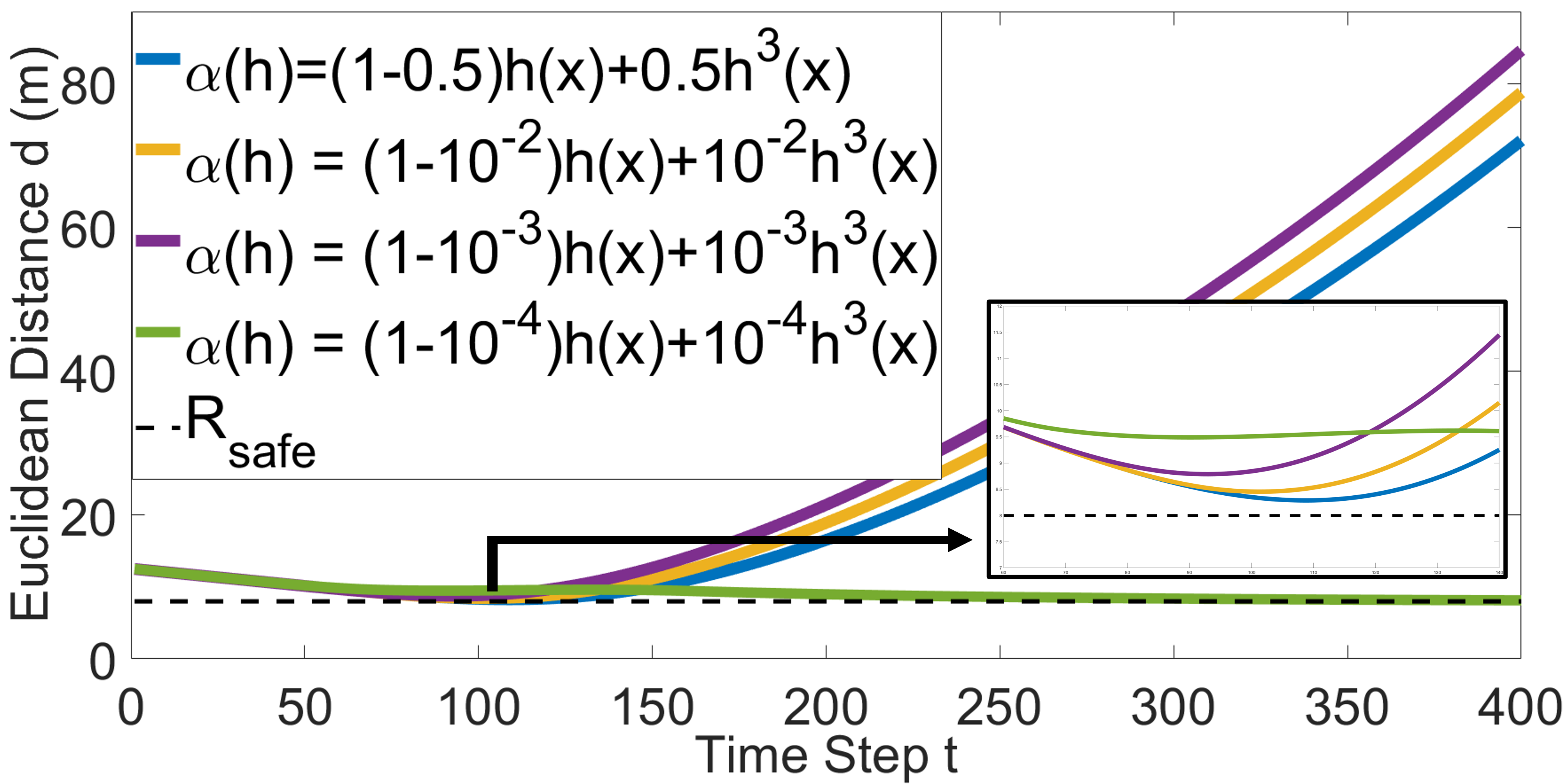}
    \caption{\footnotesize
    \label{parametricCBFcomparasion}
   Comparison of different constructions of Parametric-CBF. The total weight of the polynomial is 1, and the weights for each terms are adjusted to observe the difference. 
   }
\end{figure}

On the other hand, Parametric-CBF demonstrates its advantage of richer behavior description, as shown in Fig. \ref{parametricCBFcomparasion}. The purpose of this experiment is to reveal the effect of various relative weighting of the corresponding components on the generated behavior. Here we consider $H(x)=\begin{bmatrix}h(x) & h^3(x)\end{bmatrix}^T$. The green line has the largest relative weight (close to 1) on $h(x)$ and the smallest relative weight (close to 0) on $h^3(x)$, which makes the ego vehicle merges in the back of the merging vehicle. 
As we weigh more the higher-order components, and weigh less the first-order component, it becomes easy for the ego vehicle to merge in the front. Parametric-CBF clearly outperforms traditonal CBF with a single higher order Class K function component by achieving clearly diverse behavior, while preserving the option to expand the admissible control space as needed.

\subsection{Adaptive Behavior Improvement using Prediction}

\begin{figure}
    \centering
    \includegraphics[width = 0.8\linewidth]{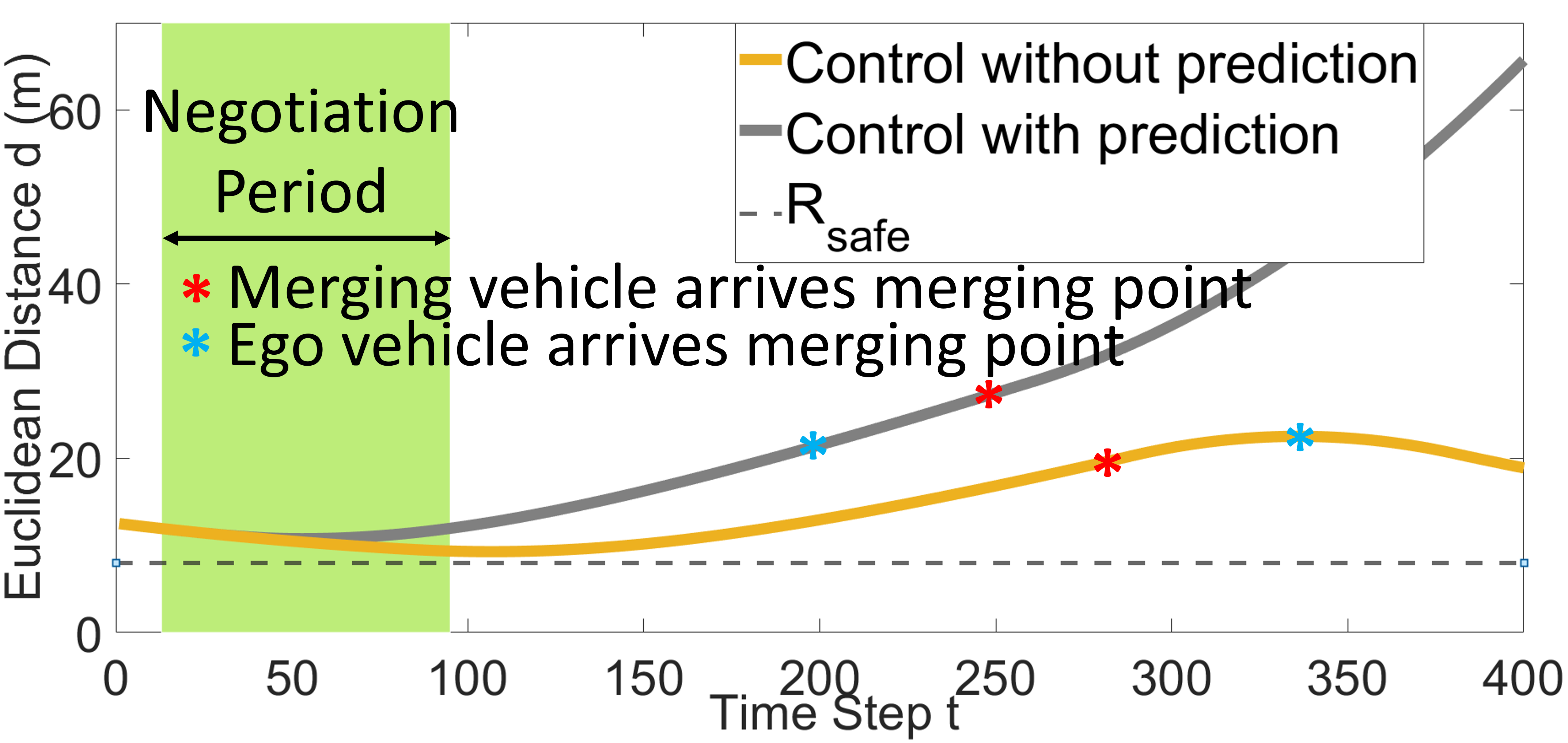}
    \caption{\footnotesize
    \label{control_prediction_loop}
   Comparison of control with prediction in the loop and without prediction. The green interval is the time period where both vehicles are negotiating with each other. The blue and red markers indicate the time step when the corresponding vehicle passes the merging point.}
\end{figure}

The goal of this application is to show how Parametric-CBF-based prediction can contribute towards more efficient safe control. The scenario we consider here is: the ego vehicle is on the main road and the merging vehicle is on the ramp. Before these two vehicles perform interactive driving, the ego vehicle gets the chance to observe the merging vehicle's behavior while it is interacting with another car. An comparison example is shown in Fig. \ref{control_prediction_loop}. The green zone indicates the time interval when the ego vehicle and the merging vehicle are negotiating, meaning their Parametric-CBF-based safety constraints are active, and the outputs of the nominal controllers are modified. 

When the ego vehicle initially started with no prediction of the merging vehicle, for safety's sake, the ego vehicle sets its Parametric-CBF $\alpha$ to be small, trying to be conservative. Therefore, the ego vehicle merges behind the merging vehicle when it does not have any prediction knowledge. Things change when the ego vehicle is able to observe the interaction between the merging vehicle and another vehicle, and makes its prediction on the merging vehicle's driving style $\alpha$. This time the ego vehicle merges in front of the merging vehicle when it leverages its prediction result and adjusts its own parameter vector $\alpha$.

Comparing the time steps when the two vehicles complete the merging task, it is observed that with the prediction involvement, the ego vehicle spent 39.6\% less time to pass the merging point, and the merging vehicle also benefits from the interaction with 10.1\% reduced time to pass the merging point. If we mark the time step when the last vehicle passes the merging point as the time completing the merging task, it is observed that by leveraging prediction results, the overall task completion time is further reduced 16.1\%. Therefore, it is obvious that Parametric-CBF-based prediction greatly contributes to improving merging task efficiency and reduce the road congestion.

\section{CONCLUSIONS}
\label{conclusion}

In this work, we propose the idea of Parametric-CBF for safe control which provides a richer behavior description while maintaining forward invariant safety. Simulation results are provided to verify the effectiveness and efficiency compared to traditional CBF. The usefulness of Parametric-CBF is demonstrated in the applications of prediction and control respectively in the autonomous driving ramp merging scenario. Heterogeneous robot interaction modeling is demonstrated by integrating prediction and control in a loop, to further improve robot working performance and efficiency. The proposed Parametric-CBF can be easily applied to other robotics applications thanks to its definition generality and computation efficiency.

\section{ACKNOWLEDGEMENTS}

The authors would like to thank Dr. Nikolay Atanasov for the stimulating discussion which brings great inspiration.

\bibliography{submission}{}
\bibliographystyle{IEEEtran}

\end{document}